\documentclass[letterpaper]{article}
\usepackage[preprint]{aaai2027}
\usepackage[hyphens]{url}
\usepackage{graphicx}
\usepackage{natbib}
\usepackage{booktabs}
\usepackage{multirow}
\usepackage{amsmath}
\usepackage{array}
\usepackage{tikz}
\newcommand{\fullmark}{\tikz[baseline=-0.55ex]\fill (0,0) circle (1.8pt);}
\newcommand{\partialmark}{\tikz[baseline=-0.55ex]{%
  \fill (0,0) -- (90:1.8pt) arc (90:270:1.8pt) -- cycle;
  \draw[line width=0.35pt] (0,0) circle (1.8pt);}}
\newcommand{\emptymark}{\tikz[baseline=-0.55ex]\draw[line width=0.35pt] (0,0) circle (1.8pt);}
\newcommand{\RouterEnFixedErnieScore}{89.43}

\newcommand{\RouterEnFixedFluxScore}{89.48}

\newcommand{\RouterEnRandomScore}{85.08}

\newcommand{\RouterEnProfileQScore}{89.72}

\newcommand{\RouterEnProfileCScore}{89.25}

\newcommand{\RouterZhFixedErnieScore}{89.58}

\newcommand{\RouterZhFixedFluxScore}{88.60}

\newcommand{\RouterZhRandomScore}{82.62}

\newcommand{\RouterZhProfileQScore}{89.98}

\newcommand{\RouterZhProfileCScore}{89.78}

\newcommand{\RouterBiFixedErnieScore}{89.51}

\newcommand{\RouterBiFixedErnieCost}{28.15}
\newcommand{\RouterBiFixedErnieSaveErnie}{0.0}
\newcommand{\RouterBiFixedErnieSaveFlux}{+55.6}

\newcommand{\RouterBiFixedFluxScore}{89.04}

\newcommand{\RouterBiFixedFluxCost}{63.34}
\newcommand{\RouterBiFixedFluxSaveErnie}{-125.0}
\newcommand{\RouterBiFixedFluxSaveFlux}{0.0}

\newcommand{\RouterBiRandomScore}{83.85}

\newcommand{\RouterBiRandomCost}{19.20}
\newcommand{\RouterBiRandomSaveErnie}{+31.8}
\newcommand{\RouterBiRandomSaveFlux}{+69.7}
\newcommand{\RouterBiProfileQScore}{89.85}

\newcommand{\RouterBiProfileQCost}{46.46}
\newcommand{\RouterBiProfileQSaveErnie}{-65.0}
\newcommand{\RouterBiProfileQSaveFlux}{+26.7}

\newcommand{\RouterBiProfileCScore}{89.51}

\newcommand{\RouterBiProfileCCost}{22.16}
\newcommand{\RouterBiProfileCSaveErnie}{+21.3}
\newcommand{\RouterBiProfileCSaveFlux}{+65.0}

\title{Scalable Question-Centric Text-to-Image Evaluation:\\Reliable Ranking, Fine-Grained Diagnosis, and Cost-Aware Routing}
\author{\normalfont\large
\textbf{Shaoan Zhao}$^{1,2,*}$, 
\textbf{Fang Zhao}$^{1,2,*}$, 
\textbf{Xueqiang Guo}$^{1,2}$, 
\textbf{Xinpei Su}$^{1,2}$,
\textbf{Huanlin Gao}$^{1,2}$, 
\textbf{Qiang Hui}$^{1,2}$, 
\textbf{Ting Lu}$^{1,2}$, 
\textbf{Fuyuan Shi}$^{1,2}$,
\textbf{Chao Tan}$^{1,2}$, 
\textbf{Bikun Yang}$^{3}$, 
\textbf{Kai Wang}$^{1,2,\dagger}$, 
\textbf{Shiguo Lian}$^{1,2,\dagger}$
\\
\textbf{ }\\[-0.1cm]
$^{1}$Data Science \& Artificial Intelligence Research Institute, China Unicom\\
$^{2}$Unicom Data Intelligence, China Unicom\\
$^{3}$China Unicom Group Co.,Ltd\\
}
\affiliations{}

\begin{document}

\maketitle
\renewcommand{\thefootnote}{}
\footnotetext[0]{%
\textsuperscript{*}Equal contribution.
\textsuperscript{$\dagger$}Corresponding author.
}
\begin{abstract}
Modern text-to-image (T2I) models often have similar total scores but different
strengths, making practical selection difficult. Fine-grained benchmarks
decompose prompts into questions, yet often return them to prompt scores and
fixed categories, weakening attribution and ignoring complexity. Related
requirements are also scored separately or as one total, obscuring basic
versus compositional failure. We present \textbf{QC-T2I-Bench}, a
question-centric framework that converts open prompts into attributed atomic
questions and organizes their dependencies with Davidsonian Scene Graphs
(DSGs). We use hierarchy-constrained question aggregation to exclude downstream
questions after a prerequisite fails and to prevent simple and complex prompts
from receiving the same total weight. We then use the DSG structure to measure
joint success within prompts and compare repeated entities across prompts,
separating basic realization failures from failures under additional
requirements. We evaluate multiple open-source T2I models on English and Chinese
prompts. The resulting question-level evidence supports reliable ranking and
fine-grained diagnosis: joint completion falls from 80.7\% for components with
two capabilities to 37.2\% for those with seven or more. Finally, we reuse the
same records for training-free routing; our cost-aware router matches ERNIE's
89.51-point estimate with 21.3\% less GPU-s/MP.
\end{abstract}

\section{Introduction}
Recent text-to-image (T2I) models have greatly improved image quality and
prompt following \cite{esser2024scaling,flux2024,wu2025qwen}. As models become
stronger, choosing between them becomes harder. Models with similar aggregate
scores may have very different capability profiles: one may render text
accurately, another may handle spatial relations reliably, and yet another may
better follow knowledge-intensive prompts. Practical model selection therefore
needs more than a leaderboard. It needs evidence about where each model
succeeds, why it fails, and which model best fits a particular request.

Conventional metrics provide global image- or prompt-level scores rather than
evidence about individual requirements \cite{hessel2021clipscore}. Evaluation
has therefore moved toward question-generation and visual-question-answering
(QG/A) methods that decompose a prompt into local checks
\cite{hu2023tifa,cho2024davidsonian,li2024genaibench}. This is an important
step, but decomposition alone does not guarantee fine-grained evaluation. The
central issue is how the resulting evidence is attributed, aggregated, and
reused.

Many evaluation pipelines still organize this evidence around prompts. They
collect prompts under predefined categories, generate local checks, average
the checks into a prompt score, and assign that score back to the prompt's
category \cite{hu2023tifa,cho2024davidsonian,wei2025tiif,
li2026qwenimagebench}. This design creates several recurring problems. A
complex prompt may contain requirements from multiple capabilities, making
its category score hard to interpret. A new request may not fit the predefined
categories. Simple and complex prompts may receive the same total weight even
though they contain different numbers of requirements. A missing parent object
may cause several dependent checks to fail, exaggerating one error. Finally,
evidence about the same requirement is rarely compared across prompts, making
it difficult to tell whether a model fails on the basic content or only after
additional constraints are added.

Table~\ref{tab:benchmark-comparison} summarizes these gaps. Atomic attribution
(A), open-prompt evaluation (O), complexity awareness (C), and cross-prompt
evidence (X) are four desired properties. Dependency awareness (D) is a
necessary safeguard against repeated penalties caused by one missing
prerequisite. Existing benchmarks support different subsets of these
properties, but none combines all five.

\begin{table}[t]
\centering
\small
\setlength{\tabcolsep}{3.6pt}
\renewcommand{\arraystretch}{1.01}
\begin{tabular*}{\columnwidth}{@{\extracolsep{\fill}}lccccc@{}}
\toprule
Benchmark & A & O & D & C & X \\
\midrule
Arena-T2I Hard   & \fullmark    & \partialmark & \fullmark    & \partialmark & \emptymark \\
ConceptMix       & \fullmark    & \partialmark & \emptymark   & \partialmark & \emptymark \\
CVTG-2K          & \partialmark & \emptymark   & \emptymark   & \partialmark & \emptymark \\
DPG-Bench        & \fullmark    & \emptymark   & \fullmark    & \emptymark   & \emptymark \\
DSG              & \fullmark    & \fullmark    & \fullmark    & \emptymark   & \emptymark \\
EvalMuse-40K     & \fullmark    & \emptymark   & \emptymark   & \emptymark   & \emptymark \\
GenAI-Bench      & \emptymark   & \fullmark    & \emptymark   & \partialmark & \emptymark \\
GenEval          & \partialmark & \emptymark   & \emptymark   & \emptymark   & \emptymark \\
GenEval 2        & \fullmark    & \emptymark   & \emptymark   & \partialmark & \emptymark \\
LongTextBench    & \partialmark & \emptymark   & \emptymark   & \partialmark & \emptymark \\
OneIG-Bench      & \partialmark & \emptymark   & \partialmark & \partialmark & \emptymark \\
PhyBench         & \partialmark & \emptymark   & \emptymark   & \emptymark   & \emptymark \\
PRISM-Bench      & \emptymark   & \emptymark   & \emptymark   & \partialmark & \emptymark \\
Qwen-Image-Bench & \fullmark    & \emptymark   & \emptymark   & \emptymark   & \emptymark \\
T2I-CompBench++  & \partialmark & \emptymark   & \emptymark   & \partialmark & \emptymark \\
T2I-CoReBench    & \fullmark    & \emptymark   & \partialmark & \partialmark & \emptymark \\
TIFA             & \fullmark    & \fullmark    & \emptymark   & \emptymark   & \emptymark \\
TIIF-Bench       & \fullmark    & \emptymark   & \emptymark   & \partialmark & \emptymark \\
WISE             & \partialmark & \emptymark   & \emptymark   & \emptymark   & \emptymark \\
\addlinespace[1.5pt]
\textbf{QC-T2I-Bench} & \fullmark & \fullmark & \fullmark & \fullmark & \fullmark \\
\bottomrule
\end{tabular*}
\caption{Support for five structural properties across T2I benchmarks.
\fullmark, \partialmark, and \emptymark\ denote full, partial, and no support.
A: atomic attribution; O: open prompts; D: dependency awareness; C: complexity
awareness; X: cross-prompt evidence.}
\label{tab:benchmark-comparison}
\end{table}

We introduce \textbf{QC-T2I-Bench}, which changes the basic unit of evaluation
from the prompt to the atomic question. Each record preserves the question's
capability label, visual answer, dependency validity, and prompt context. A
two-level taxonomy organizes questions into 21 capabilities under Non-text
Entities, Text Content, Attributes, Relations, and High-Level Semantics. Hierarchy-Constrained
Question Aggregation (HCQ) removes dependency-invalid evidence and balances
capabilities instead of giving every prompt the same total weight. Because
prompts use a shared question schema rather than benchmark-specific prompt
categories, the same protocol can also be applied to new requests.

We further use DSGs to organize the questions in each prompt according to
their dependencies \cite{cho2024davidsonian}. This lets us test whether a
model can satisfy a group of related requirements together, rather than only
checking them one by one. We then compare repeated entities across different
prompts to distinguish a basic generation failure from a failure caused by
additional attributes or relations. The same question records also form a
capability profile for each model and can guide generator selection for new
requests.

We evaluate QC-T2I-Bench on 13 T2I models with bilingual evidence. The
framework supports three connected uses: \textbf{reliable ranking},
\textbf{fine-grained diagnosis}, and \textbf{cost-aware routing}. Bootstrap
analysis separates most model pairs while preserving uncertainty for the
closest systems. The DSG analysis shows where models fail as related
requirements accumulate. Finally, a training-free cost-aware router matches
the fixed ERNIE point estimate while reducing GPU-s/MP by 21.3\%.

Our contributions are threefold. \textbf{(1)} We introduce QC-T2I-Bench,
which converts open prompts into attributed atomic questions and uses
validity-aware Hierarchy-Constrained Question Aggregation (HCQ) to avoid
prompt-level normalization. \textbf{(2)} We develop a DSG-based compositional
analysis that combines complete-component success, cross-prompt root controls,
and topology-matched contrasts to localize failures. \textbf{(3)} We
demonstrate reliable ranking, fine-grained diagnosis, and training-free
quality--cost routing, including a 21.3\% cost reduction at the matched ERNIE
point estimate.

\section{Related Work}
\paragraph{Fine-grained evaluation benchmarks.}
T2I evaluation has progressed from controlled property tests to question-based
verification. GenEval \cite{ghosh2023geneval} evaluates predefined object
properties, while GenEval~2 \cite{kamath2025geneval} adds atom-level questions
and atomicity analysis. TIFA \cite{hu2023tifa} decomposes prompts into VQA
pairs, and DSG \cite{cho2024davidsonian} organizes atomic questions through
valid dependencies. Other work improves supervision and calibration through
dimension-specific concepts or element-level annotations
\cite{wei2025tiif,han2024evalmuse40kreliablefinegrainedbenchmark}, while
VQAScore and GenAI-Bench study holistic alignment and its agreement with human
preferences \cite{lin2024evaluating,li2024genaibench}. More recent benchmarks
use hierarchical capability rubrics or dependency-aware checklists
\cite{li2026qwenimagebench,ban2026arenat2ihard}. These developments make local
verification more reliable, but their evidence is usually summarized within
each prompt or a fixed reporting taxonomy.

\paragraph{Capability-specific and closed-set benchmarks.}
A complementary line of work deepens evaluation within selected capabilities.
WISE and PhyBench focus on world knowledge and physical rules
\cite{niu2025wise,meng2024phybench}; T2I-CompBench++, ConceptMix, and DPG-Bench
stress compositional or dense instructions
\cite{huang2023t2i,wu2024conceptmix,hu2024ella}; and PRISM-Bench, OneIG-Bench,
and T2I-CoReBench expand coverage to bilingual, creative, and reasoning-heavy
scenarios \cite{fang2025flux,chang2026oneig,li2025easier}. LongTextBench and
CVTG-2K further examine long or multilingual instructions and visual text
\cite{geng2025x,du2025textcrafter}. Table~\ref{tab:benchmark-comparison}
summarizes these structural choices: atomic checking is common, but support for
open prompts, dependencies, and prompt complexity is fragmented; none of the
compared benchmarks connects evidence across prompts. QC-T2I-Bench combines
all five properties in one question-centric framework.

\paragraph{Evidence reuse and generator routing.}
Generator-routing work addresses model selection directly. CATImage learns
prompt-conditioned quality--cost decisions, while DiffAgent uses an LLM agent
for API selection \cite{li2025cost,zhao2024diffagent}. OctoT2I, Image-POSER,
and GenArtist extend selection to stateful or multi-step orchestration through
self-evolving memory, reinforcement learning, or tool planning
\cite{jiang2026octot2i,mohebbi2025image,wang2024genartist}. These methods obtain
routing signals by learning a policy, building agent state, or repeatedly
evaluating intermediate outputs. Our router instead reuses question-level
histories already collected for ranking and diagnosis, enabling training-free
model selection from the same evaluation evidence.

\section{Method}
QC-T2I-Bench retains the identity, capability coordinates, and dependency
context of every atomic judgment. These records support two complementary
operations. The scoring path aggregates valid non-text and text evidence into
capability coordinates and a hierarchy-constrained question aggregation (HCQ)
score. The structural path
reuses the recorded DSGs to measure component completion, construct
cross-prompt root controls, and test topology-localized outcome coupling. The
same records subsequently support model diagnosis and training-free routing.

\paragraph{Question construction and capability coordinates.}
We collect prompts from six public sources spanning knowledge, entities,
attributes, relations, text rendering, reasoning, and long multilingual
instructions; source counts are provided in the supplement. Each prompt $p$ is converted into
$c(p)=\{(q_i,t_i)\}_{i=1}^{n_p}$, where $n_p$ is the number of atomic questions,
$q_i$ is an independently judgeable question, and $t_i$ is its secondary
capability label. Construction follows four rules: \textbf{Target Yes},
\textbf{Atomicity}, \textbf{Coverage}, and \textbf{Tag Validity}. Together,
they require a ``yes'' answer for a compliant image, one visual requirement per
question, coverage of salient constraints, and attribution by the question's
core predicate. We instantiate this fixed contract with
Qwen3-235B-A22B \cite{yang2025qwen3}, iterative automated auditing, and a final
cleanup pass; the supplement reports source composition and audit details.

\begin{figure}[t]
\centering
\includegraphics[width=.96\columnwidth]{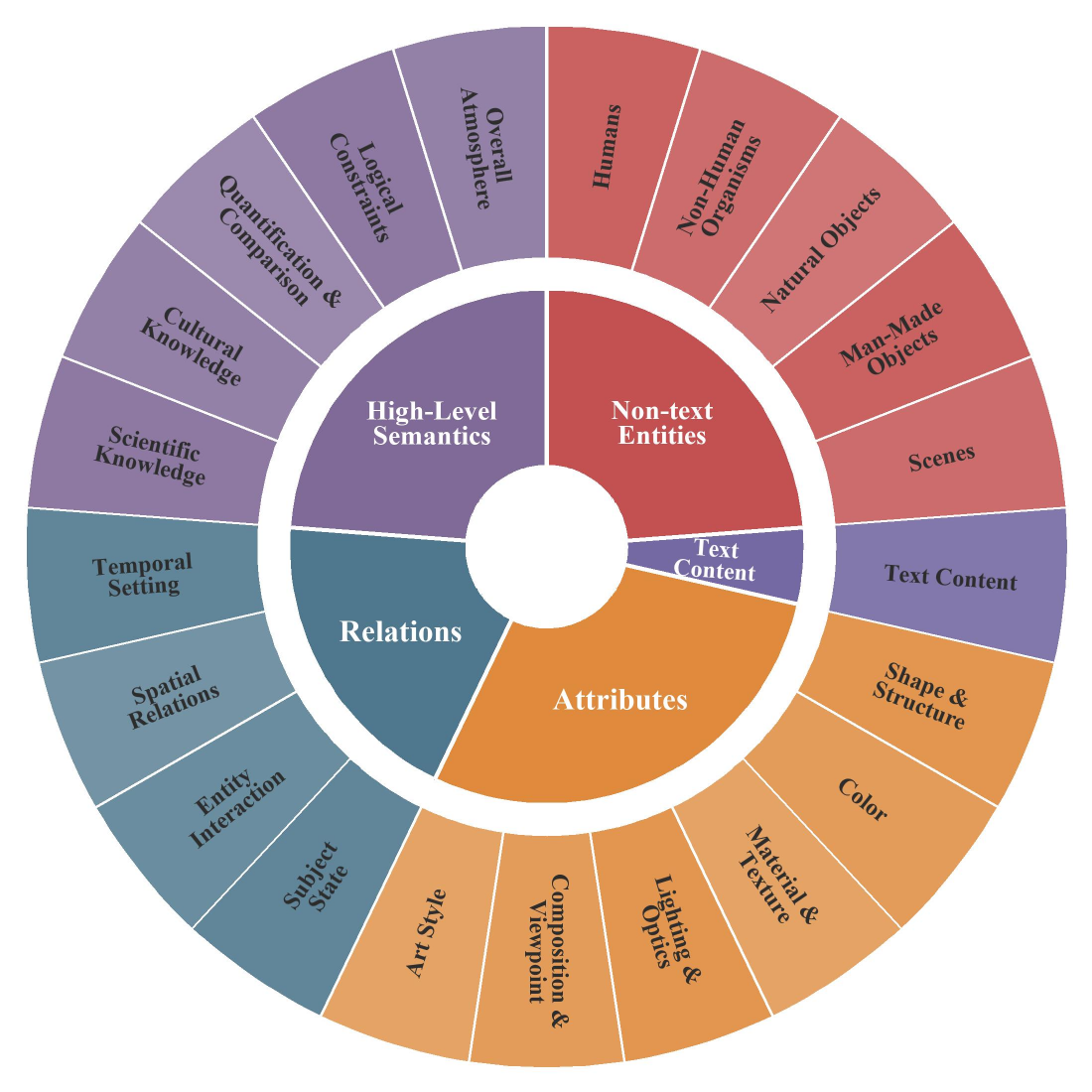}
\caption{Two-level taxonomy: five first-level reporting groups organize 21 mutually exclusive secondary capabilities, assigned by each question's core predicate.}
\label{fig:taxonomy}
\end{figure}

Evaluation produces one record per question,
\begin{equation}
r=(p,q,m,\ell,g,t,y,v),
\end{equation}
where $p$ and $q$ are the prompt and question, $m$ is the evaluated model,
$\ell$ is the language, $g$ is the first-level reporting group, and $t$ is the
secondary capability. For non-text capabilities, $y\in\{0,1\}$ is the native
binary outcome and $v\in\{0,1\}$ indicates dependency validity. Text Content
retains the same capability coordinates but uses the transcription statistics
defined below for official scoring.
The taxonomy contains five first-level groups---\emph{Non-text Entities},
\emph{Text Content}, \emph{Attributes}, \emph{Relations}, and
\emph{High-Level Semantics}---and 21 secondary capabilities
(Figure~\ref{fig:taxonomy}). Each question receives exactly one secondary
capability by its core predicate, while one prompt may contribute questions to
several groups. Text Content is separated because it uses the dedicated
transcription evaluator described below.

\paragraph{Dependency-aware capability scoring.}
Before adjudication, we organize prompt $p$ into a Davidsonian Scene Graph
$G_p=(V_p,E_p)$ \cite{cho2024davidsonian}, where $V_p$ is the set of question
nodes and $E_p$ is the set of directed prerequisite edges. An edge
$u\!\to\!q\in E_p$ indicates that question $u$ is a semantic prerequisite of
$q$. During adjudication, $q$ is scored only when its prerequisites succeed,
preventing one missing entity from becoming repeated attribute and relation
failures.

For model
$m$, language scope $\Lambda$, and non-text capability $t$, let
$\mathcal Q_{\ell,t}$ be the questions in language $\ell$,
$y_{m\ell q}\in\{0,1\}$ their adjudicated outcomes, and
$v_{m\ell q}\in\{0,1\}$ their dependency validity. The valid count and score are
\begin{equation}
\begin{aligned}
N_{m,\Lambda,t}
&=\sum_{\ell\in\Lambda}\sum_{q\in\mathcal{Q}_{\ell,t}}v_{m\ell q},\\[-1mm]
S_{m,\Lambda,t}
&=\frac{1}{N_{m,\Lambda,t}}
\sum_{\ell\in\Lambda}\sum_{q\in\mathcal{Q}_{\ell,t}}
v_{m\ell q}y_{m\ell q},\quad t\ne t_{\mathrm{text}}.
\end{aligned}
\label{eq:secondary-binary-score}
\end{equation}
Thus $S_{m,\Lambda,t}$ is the mean over valid questions at one capability
coordinate when $N_{m,\Lambda,t}>0$ and is NA otherwise; invalid records enter
neither sum.

\paragraph{Text-rendering branch.}
Rendered text remains a semantic capability coordinate, but its evidence type
differs: a binary VQA judgment can accept semantically related text while
missing character-level errors. We therefore use Qwen3-VL-Instruct-30B to
extract text blocks and optimally match them to targets within each image $i$.
Let $D_{m\ell i}$ be the target-conditioned UTF-16 edit distance and
$L_{\ell i}$ the target length, and let $\mathcal I^{\mathrm{text}}_{m\ell}$
contain images with an available transcription record. We micro-average over
all available text records in language scope $\Lambda$:
\begin{equation}
S_{m,\Lambda,t_{\mathrm{text}}}
=\max\!\left(0,1-
\frac{\sum_{\ell\in\Lambda}\sum_{i\in\mathcal I^{\mathrm{text}}_{m\ell}}D_{m\ell i}}
{\sum_{\ell\in\Lambda}\sum_{i\in\mathcal I^{\mathrm{text}}_{m\ell}}L_{\ell i}}\right).
\label{eq:text-global-score}
\end{equation}
This micro-average gives every target character equal weight; empty extractions
incur full deletion cost, while unavailable evaluator records are excluded. If
the denominator is zero, the score is NA.

\paragraph{Hierarchy-constrained aggregation.}
We combine these secondary scores with HCQ, which micro-averages valid questions within each secondary
capability, then macro-averages capabilities within each reporting group and
across the five groups. With $\mathcal{T}_g$ the capabilities in group $g$,
$T_g=|\mathcal{T}_g|$, and $G$ the number of reporting groups,
\begin{equation}
S_{m,\Lambda}^{\mathrm{HCQ}}
=\frac{1}{G}\sum_{g=1}^{G}\frac{1}{T_g}
\sum_{t\in\mathcal{T}_g}S_{m,\Lambda,t}.
\label{eq:hcq-overall}
\end{equation}
Here $G=5$ and, in the order Non-text Entities, Text Content, Attributes,
Relations, and High-Level Semantics, $(T_1,\ldots,T_5)=(5,1,6,4,5)$.

\paragraph{Compositional capability analysis.}
Beyond capability aggregation, we use the recorded DSG structure for three
complementary diagnostics. Component exactness tests whether related
requirements are jointly satisfied; cross-prompt root matching separates
entity-generation difficulty from failures under attached constraints; and
marginal-controlled coupling tests whether excess outcome dependence localizes
to explicit DSG edges.

A prompt may contain several disconnected structures, so we define its
\emph{DSG atlas} as the maximal weakly connected components
$\mathcal A_p=\{G_{p,1},\ldots,G_{p,K_p}\}$. For a component $G$ of a fixed
prompt--language pair, let $\mathcal Q(G)$ be its questions and
$\mathcal T(G)=\bigcup_{q\in\mathcal Q(G)}\{t(q)\}$ its component tag set. We
score each maximal component once for its full tag set rather than enumerating
its pairs or lower-order subsets. Components define structural context, not an
assumption of statistical dependence; that distinction motivates the
edge-controlled test below.

For model $m$ and component $G$, $C_m(G)$ is the complete-component exactness
indicator. It equals 1 only when every question is valid and succeeds, equals 0
when any valid question fails, and is undefined when the component cannot
otherwise be scored. Unlike mean coverage, this criterion retains root failures
even when their descendants are masked:
\begin{equation}
C_m(G)=
\begin{cases}
0,&\exists q\in\mathcal Q(G):\ v_{mq}=1\land y_{mq}=0,\\
1,&\forall q\in\mathcal Q(G):\ v_{mq}=1\land y_{mq}=1,\\
\mathrm{NA},&\text{otherwise.}
\end{cases}
\label{eq:complete-component-exact}
\end{equation}
Applying the same rule separately to roots and descendants gives root survival
$R_m(G)$ and descendant completion conditional on root survival. To distinguish
a composition-specific root failure from a generally difficult entity, we
conservatively match entity roots across prompts without adding cross-prompt
dependency edges. For entity identity $e$ in prompt $p$, the leave-one-prompt-out
baseline is
\begin{equation}
\bar R^{-p}_{m\ell e}=\frac{1}{|\mathcal P_{\ell e}\setminus\{p\}|}
\sum_{p'\in\mathcal P_{\ell e}\setminus\{p\}}R_{m\ell p'e}.
\label{eq:cross-prompt-root}
\end{equation}
Here, $\mathcal P_{\ell e}$ contains prompts in language $\ell$ with a matched
root $e$, and $R_{m\ell p'e}\in\{0,1\}$ is its prompt-level root-survival
outcome for model $m$. Matches stay within one language and model and require
at least three other prompts. Comparing the local outcome $R_{m\ell pe}$ with
$\bar R^{-p}_{m\ell e}$ therefore controls for the model's baseline ability to
realize the same entity.

\paragraph{Marginal-controlled coupling.}
Complete-component exactness decreases as more fallible requirements are
conjoined even when their outcomes are independent. We therefore test whether
joint success follows recorded DSG topology beyond this ordinary multiplication
of marginal success rates. For model $m$, language $\ell$, prompt $p$, matched
pair stratum $s$, and context $c$, define
\begin{equation}
\begin{aligned}
\widehat P^{c,11}_{m\ell ps}
&=\frac{1}{|\mathcal M_{\ell psc}|}
  \sum_{(i,j)\in\mathcal M_{\ell psc}}
  Y_{m\ell pi}Y_{m\ell pj},\\
\Delta^{c}_{m\ell ps}
&=\widehat P^{c,11}_{m\ell ps}
-\hat\pi^{-p,c}_{m\ell s,1}\hat\pi^{-p,c}_{m\ell s,2}.
\end{aligned}
\label{eq:dependency-residual}
\end{equation}
Here the nonempty set $\mathcal M_{\ell psc}$ contains ordered matched pairs
$(i,j)$, $Y_{m\ell pi},Y_{m\ell pj}\in\{0,1\}$ are their native binary
outcomes, and positions 1 and 2 denote the two ordered endpoints. The stratum
fixes their ordered capability tags, root/descendant roles, and DSG depths;
$c$ is either \emph{edge}, a direct parent--child pair, or \emph{disc}, a
structurally matched pair from disconnected components in the same prompt.
The first term is observed joint success, while the product of
leave-one-prompt-out endpoint marginals is the success expected in the same
model--language stratum.

To isolate dependence associated specifically with a recorded DSG edge, we
compare the two residuals after matching:
\begin{equation}
\Gamma_{m\ell}=\frac{1}{|\mathcal P^*_{m\ell}|}
\sum_{p\in\mathcal P^*_{m\ell}}
\frac{1}{|\mathcal S^*_{m\ell p}|}
\sum_{s\in\mathcal S^*_{m\ell p}}
\left(\Delta^{\mathrm{edge}}_{m\ell ps}
-\Delta^{\mathrm{disc}}_{m\ell ps}\right).
\label{eq:dependency-contrast}
\end{equation}
Here $\mathcal P^*_{m\ell}$ contains prompts with matched edge and disconnected
contexts, and $\mathcal S^*_{m\ell p}$ contains their eligible matched strata;
the nested means give equal weight to strata within a prompt and then to
prompts. We additionally macro-average these estimates across models. Thus
$\Delta=0$ is consistent with marginal multiplication, while $\Delta>0$
indicates excess outcome coupling. The contrast $\Gamma$ asks whether that
excess is stronger on explicit DSG edges than on matched disconnected pairs;
it is not a quality gain or a causal composition penalty. For this structural
test only, $Y$ is recovered before dependency masking, which would otherwise
induce edge dependence mechanically. Matching thresholds and full estimators
are reported in the supplement.

Together, the scoring and structural paths populate the evidence summarized by
the bilingual views in Table~\ref{tab:primary}; the following experiments reuse
the same records for ranking, diagnosis, and routing.

\begin{table*}[t]
\centering
\small
\setlength{\tabcolsep}{9.0pt}
\renewcommand{\arraystretch}{1}
\begin{tabular}{@{}l*{6}{r}@{\hspace{7pt}}*{6}{r}@{}}
\toprule
Model & \multicolumn{6}{c}{English} & \multicolumn{6}{c}{Chinese} \\
\cmidrule(lr){2-7}\cmidrule(lr){8-13}
& Total & Ent. & Text & Attr. & Rel. & Sem. & Total & Ent. & Text & Attr. & Rel. & Sem. \\
\midrule
FLUX.2-dev & \textbf{89.48} & \textbf{97.5} & 78.6 & 94.7 & 89.4 & \textbf{87.1} & 88.60 & 97.2 & 74.4 & 94.6 & 89.6 & \textbf{87.3} \\
ERNIE-Image & 89.43 & 97.2 & \textbf{79.7} & \textbf{95.0} & \textbf{89.5} & 85.7 & \textbf{89.58} & 97.5 & 79.7 & \textbf{95.4} & \textbf{90.5} & 84.8 \\
Z-Image Base & 88.11 & 95.5 & 79.1 & 93.6 & 86.9 & 85.3 & 89.15 & 96.7 & 79.8 & 94.3 & 88.5 & 86.4 \\
Qwen-Image & 87.72 & 95.7 & 79.4 & 93.4 & 87.0 & 83.1 & 88.55 & 96.3 & \textbf{82.4} & 93.6 & 87.9 & 82.6 \\
Lens & 87.34 & 97.3 & 72.8 & 94.6 & 89.1 & 82.9 & 87.37 & \textbf{97.5} & 73.0 & 94.4 & 89.6 & 82.3 \\
FLUX.2-Klein-9B & 87.08 & 95.8 & 72.3 & 94.0 & 87.9 & 85.3 & 85.35 & 96.7 & 61.0 & 94.4 & 89.1 & 85.5 \\
HiDream-O1 & 86.80 & 95.3 & 73.1 & 92.7 & 87.3 & 85.5 & 87.00 & 95.1 & 76.1 & 92.2 & 87.0 & 84.6 \\
Z-Image Turbo & 85.67 & 93.9 & 78.5 & 91.6 & 84.0 & 80.4 & 85.98 & 95.1 & 77.5 & 91.4 & 84.8 & 81.1 \\
FLUX.2-Klein-4B & 84.33 & 96.0 & 61.9 & 93.5 & 86.8 & 83.4 & 82.29 & 95.3 & 52.0 & 93.6 & 88.3 & 82.3 \\
GLM-Image & 83.94 & 92.6 & 73.4 & 88.9 & 82.9 & 82.0 & 84.88 & 93.1 & 76.2 & 89.1 & 83.9 & 82.2 \\
LongCat-Image & 83.91 & 96.2 & 63.8 & 91.6 & 84.7 & 83.3 & 85.97 & 96.2 & 70.0 & 92.8 & 87.7 & 83.2 \\
FLUX.1-dev & 76.21 & 89.9 & 44.1 & 88.8 & 80.1 & 78.2 & 32.75 & 23.0 & 16.5 & 47.2 & 41.8 & 35.3 \\
HiDream-I1 & 75.66 & 91.7 & 35.2 & 88.9 & 81.7 & 80.8 & 65.02 & 78.9 & 20.6 & 80.4 & 73.8 & 71.4 \\
\bottomrule
\end{tabular}
\caption{Aligned English and Chinese HCQ results (\%). Models are ordered by English total; each language is aggregated independently by equally averaging the five reporting views. Ent., Attr., Rel., and Sem. denote Non-text Entities, Attributes, Relations, and High-Level Semantics. Best values within each language are bold; full 21-dimension profiles are in the supplement.}
\label{tab:primary}
\end{table*}

\section{Experiments}
\subsection{Experimental Setup}
\paragraph{Models and generation.}
We evaluate 13 recent open-source T2I systems: ERNIE-Image
\cite{liu2026ernieimagetechnicalreport}, Qwen-Image \cite{wu2025qwen}, Z-Image
Base and Turbo \cite{zimage2025}, FLUX.1-dev and FLUX.2-dev \cite{flux2024}, FLUX.2-Klein
4B/9B, GLM-Image \cite{zai2026glmimage}, HiDream-O1/I1
\cite{cai2026hidreamo1,cai2025hidreami1}, LongCat-Image \cite{longcat2025}, and
Lens \cite{chen2026lens}. We use official defaults and disable configurable
prompt enhancement so every generator receives the same prompt.

The fixed benchmark contains 6,573 conceptual prompts, 94,547 English
questions, and 94,555 Chinese questions. Analyses are run separately by
language: the main paper presents both leaderboards side by side and reports bilingual
aggregation, dependency, and routing summaries, while complete
language-specific profiles are in the supplement.

\paragraph{Evaluation protocol.}
Every model generates both language sets. We apply HCQ
(Equation~\ref{eq:hcq-overall}) independently to each language; the
language-specific leaderboards do not pool English and Chinese evidence.

\paragraph{Aggregation robustness.}
Rank recovery alone confounds scoring noise with rule-induced changes in model
gaps. Let $g(t)$ map capability $t$ to its unique reporting group. We hold the
HCQ capability weight $\alpha_t=1/(G T_{g(t)})$ fixed and compare the noise introduced by two
within-capability rules. Let $\mathcal P_t$ be the prompts contributing to
capability $t$, $P_t=|\mathcal P_t|$, $n_{pt}$ the valid question count from
prompt $p$, and $N_t=\sum_{p\in\mathcal P_t}n_{pt}$. Question-micro weighting assigns each judgment
$\alpha_t/N_t$; prompt-first weighting assigns it
$\alpha_t/(P_tn_{pt})$. Under independent, zero-mean, equal-variance
adjudication errors, their noise-variance ratio is
\begin{equation}
\frac{V_t^{\mathrm{prompt}}}{V_t^{\mathrm{question}}}
=\left(\frac{1}{P_t}\sum_{p\in\mathcal P_t}n_{pt}\right)
  \left(\frac{1}{P_t}\sum_{p\in\mathcal P_t}\frac{1}{n_{pt}}\right)\ge 1.
\label{eq:aggregation-noise-variance}
\end{equation}
Here $V_t^{\mathrm{prompt}}$ and $V_t^{\mathrm{question}}$ are the variances
of capability $t$'s weighted adjudication-error contribution under the two
rules.
The arithmetic--harmonic mean inequality makes the ratio at least one, with
equality only when every prompt contributes the same number of questions.
Thus equal atomic weights minimize variance under the fixed capability
construct, while prompt-first weighting amplifies errors in relatively sparse
prompts. The full proof, unequal-variance extension, and rank-gap analysis are
reported in the supplement.

\subsection{Reliable Ranking}
\paragraph{Leaderboard.}
Table~\ref{tab:primary} aligns the five capability views and HCQ total across
languages; complete 21-dimensional profiles are in the supplement. FLUX.2-dev
leads in English and ERNIE-Image in Chinese, but no model is best across every
view, keeping each scalar order traceable to its diagnostic coordinates.

\paragraph{Bootstrap ranking reliability.}
We test whether the leaderboard survives prompt resampling using 2,000 paired
bootstrap draws over the full set of retained conceptual-prompt clusters,
resampling every attached question together. The 95\% paired intervals exclude
zero for 68 of the 78 model pairs. FLUX.2-dev has a 55.2\% top-1 probability
and a 95\% rank interval of $[1,2]$; ERNIE-Image has the complementary 44.8\%
top-1 probability and the same interval. Thus the broad ordering is resolved,
but the 0.042-point gap between the first two models is not evidence of
deterministic separation.

\paragraph{Adjudication-noise robustness.}
Prompt-first normalization increases both adjudication-noise variance and the
influence of one error in a short prompt in English and Chinese
(Table~\ref{tab:main-adjudication-noise}). The smaller full-benchmark effects
would hide this difference at the roughly 1,000-prompt scale of recent
benchmarks; complete diagnostics are in the supplement.

\begin{table}[t]
\centering
\small
\setlength{\tabcolsep}{2.0pt}
\begin{tabular}{lrrrr}
\toprule
& Var. increase & \multicolumn{3}{c}{One short-prompt error (pp)} \\
\cmidrule(lr){3-5}
Language & Prompt vs. Q & Q & Prompt & Increase \\
\midrule
English & 23.2\% & 0.0102 & 0.0151 & 48.4\% \\
Chinese & 21.2\% & 0.0110 & 0.0156 & 41.8\% \\
\bottomrule
\end{tabular}
\caption{Adjudication-noise sensitivity at 1,000 prompts. ``Question'' gives
equal weight to valid atomic judgments; ``Prompt'' normalizes prompts first. A
short prompt contains at most three questions. Lower is more robust.}
\label{tab:main-adjudication-noise}
\end{table}

\begin{figure*}[!t]
\centering
\includegraphics[width=.98\textwidth]{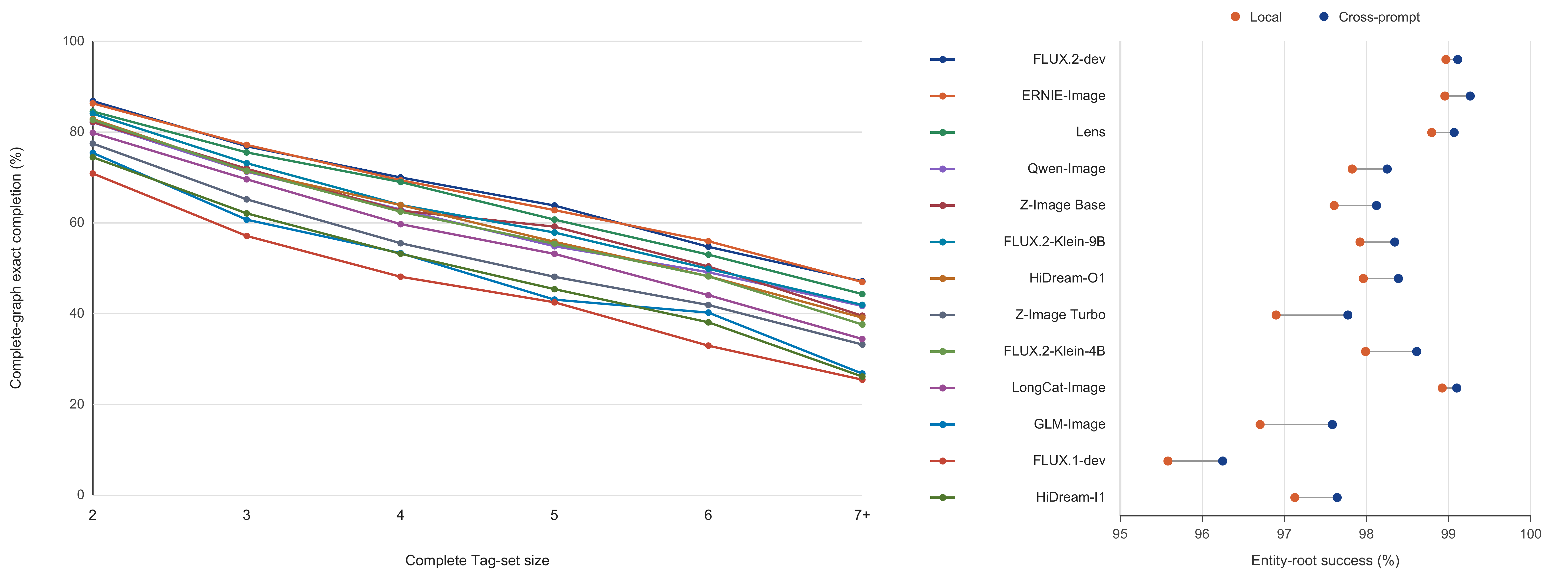}
\caption{Scaling and root-control views of complete DSG components.
Left: prompt-macro exact completion versus the number of distinct capability tags in the
full graph. Right: local entity-root success (orange) and the same model's
leave-one-prompt-out success on matched roots (blue). Line samples identify the
corresponding model curves.}
\label{fig:composition-scale}
\end{figure*}

\begin{figure*}[!t]
\centering
\includegraphics[width=.78\textwidth]{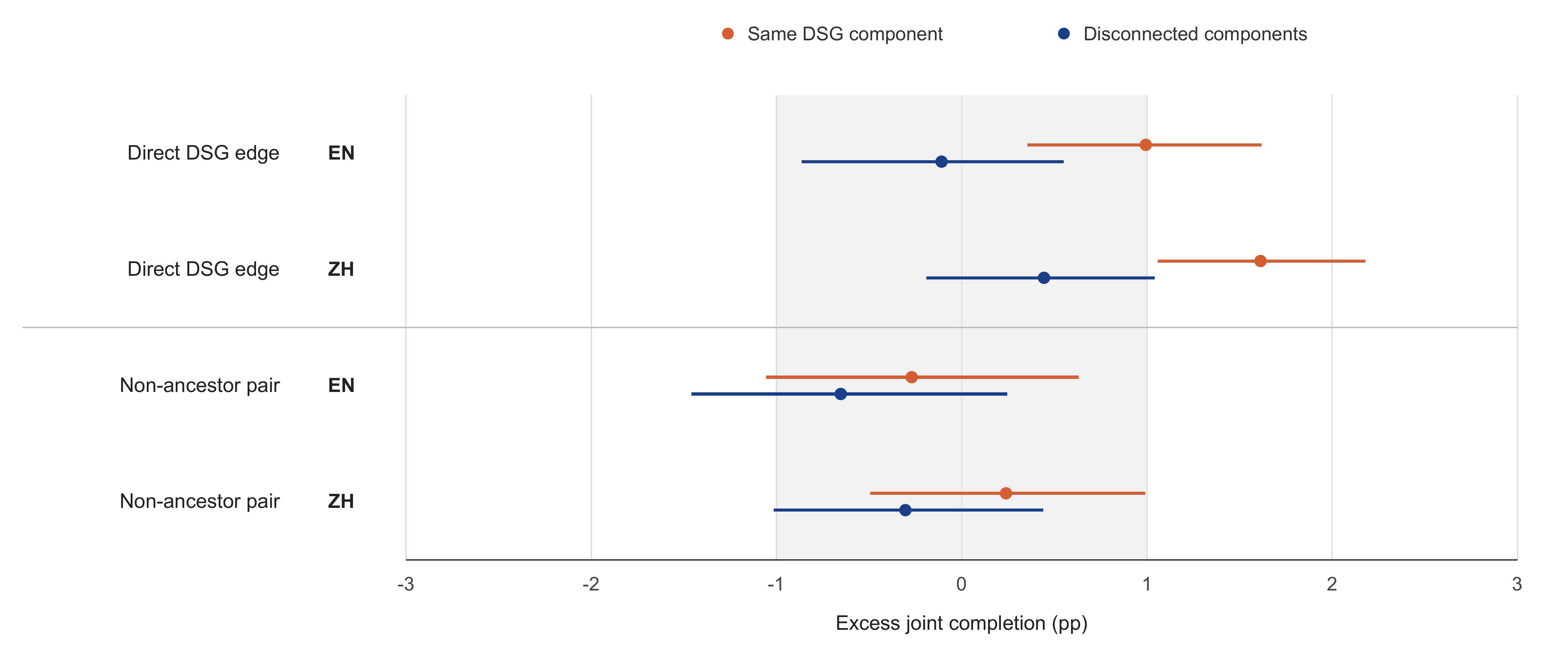}
\caption{Outcome coupling after removing the joint success expected from
ordinary marginal error multiplication. Each row compares question pairs in
the same DSG component (orange) with matched pairs from disconnected components
(blue). Points are model-macro residuals $\Delta$ from
Equation~\ref{eq:dependency-residual}; bars are 95\% prompt-bootstrap intervals.}
\label{fig:dependency-coupling}
\end{figure*}

\begin{figure*}[!t]
\centering
\includegraphics[width=.94\textwidth]{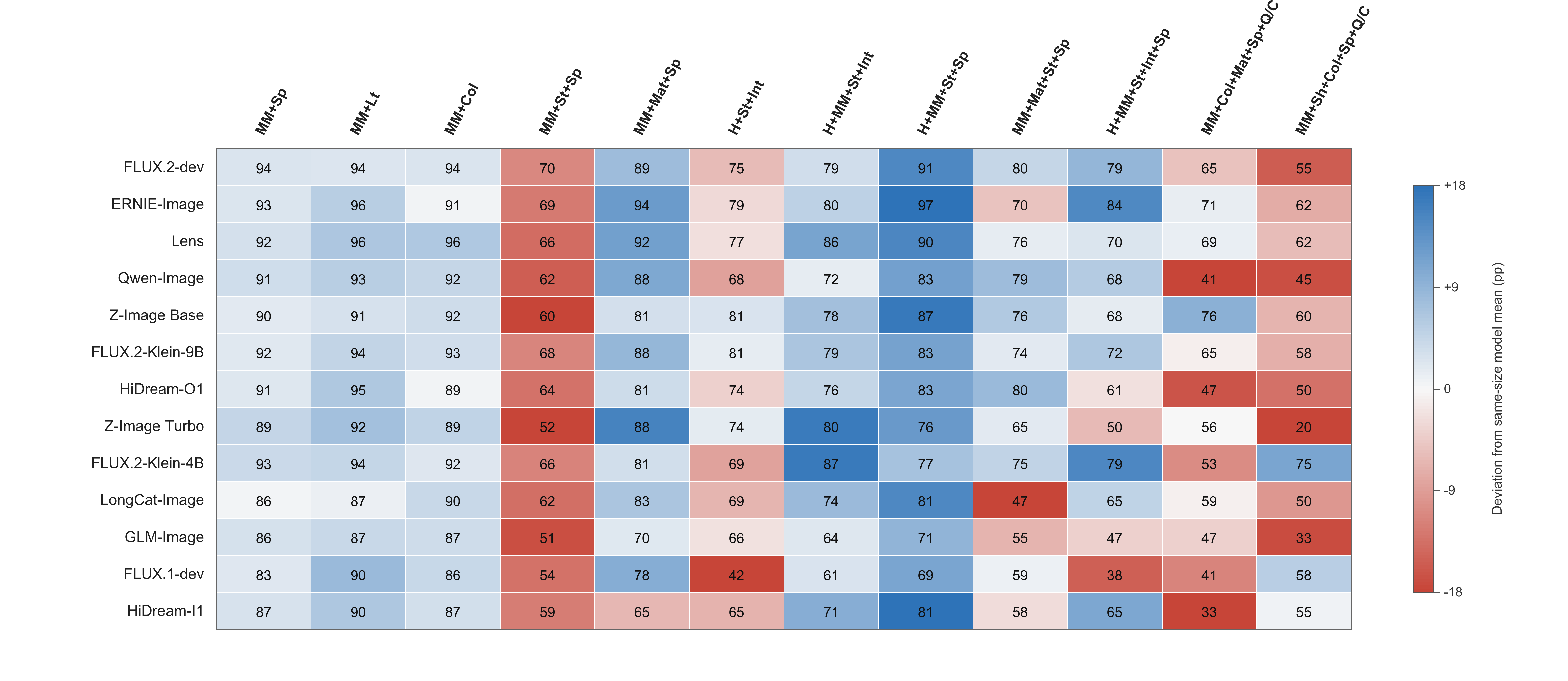}
\caption{High-frequency DSG-component composition fingerprints.
Columns are the three most frequent full tag sets at each cardinality from two
through five; frequency selects columns but does not weight cells. Abbreviations:
H=Humans, MM=Man-Made, Sh=Shape, Col=Color, Mat=Material, Lt=Lighting,
St=State, Int=Interaction, Sp=Spatial, and Q/C=Quantification/Comparison.
Cell text is prompt-macro conditional descendant completion (\%) after every
root survives. The color bar gives the deviation in percentage points from the
same model's mean at that tag-set size.}
\label{fig:full-tagset-fingerprint}
\end{figure*}

\subsection{Fine-Grained Diagnosis}
\paragraph{Capability boundaries.}
The five views in Table~\ref{tab:primary} separate strong entity and appearance
generation from weaker text, relation, and semantic capabilities. The complete
21-dimensional matrices further localize persistent deficits in Text Content,
Scientific and Cultural Knowledge, Temporal Setting, and Logical Constraints;
cross-language shifts are capability- and model-specific rather than uniform.

\paragraph{Effect of dependency masking.}
The official metrics exclude dependency-invalid descendants. Counting them as
failures leaves the broad ordering stable ($\rho=0.978$) but changes two
positions, indicating that masking removes cascading penalties without
manufacturing the ranking; full shifts are in the supplement.

\paragraph{DSG-component composition diagnosis.}
The DSG atlas yields 10,584 maximal components from 5,792 conceptual prompts.
In Figure~\ref{fig:composition-scale}, mean exact completion falls from 80.7\%
for two-tag components to 37.2\% for components with seven or more tags. This
describes joint difficulty, not interaction: exact success becomes stricter as
requirements accumulate even under independent errors.

The cross-prompt control uses 4,319--4,347 eligible matched contexts per model. Local
roots trail their matched baseline by only 0.14--0.88 points, with nine of
thirteen intervals excluding zero. Root survival and conditional descendant
completion can therefore locate entity versus attached-structure failure, but
do not by themselves establish interaction.

\paragraph{Topology-localized coupling.}
After marginal control, the direct-edge contrast is $+1.10$ points (95\%
prompt-bootstrap interval $[+0.69,+1.50]$), whereas the non-ancestral contrast
is smaller and unresolved at $+0.38$ points $[-0.17,+1.01]$
(Figure~\ref{fig:dependency-coupling}). Directly dependent requirements thus
co-succeed and co-fail beyond their individual rates, and the excess localizes
to explicit DSG topology rather than component membership alone. Chinese
results reproduce the direct-edge pattern; full estimates are in the supplement.

Figure~\ref{fig:full-tagset-fingerprint} adds structure-specific resolution:
Qwen-Image is weak on a five-tag Man-Made--Color--Material--Spatial--Quantification
component, whereas FLUX.2-Klein-4B is comparatively strong on the related
Shape variant. Such profiles distinguish models with similar scalar scores;
all cells and structural frequencies are in the supplement.

\subsection{Scalable Cost-Aware Routing}
Question histories reveal which generators satisfy which atomic requirements.
We reuse them to select one T2I generator for each complete request; routing
does not select the VQA evaluator.

\paragraph{Routers and evaluation protocol.}
Our training-free router, \emph{Q-Profile}, estimates generator quality from
the request's active views and reference-fold outcomes without learned parameters. Its
quality-first variant (Q) selects the predicted maximum; its cost-aware variant
(C) selects the least expensive generator within a validation-selected
$\epsilon$ of that maximum.

We retain the original source-stratified nested five-fold choices and do not
retune settings after the HCQ revision. English and Chinese scores are computed
separately and combined with fixed 50/50 weights. Cost is latency times occupied
GPUs per output megapixel (GPU-s/MP); paired intervals resample shared conceptual
IDs across languages. Full protocol and retrieval ablations are in the supplement.

\begin{table}[t]
\centering
\small
\setlength{\tabcolsep}{1.0pt}
\renewcommand{\arraystretch}{1.5}
\begin{tabular}{lrrrrr}
\toprule
Method & EN & ZH & Bi. & GPU-s/MP & Save E/F \\
\midrule
ERNIE-Image & \RouterEnFixedErnieScore & \RouterZhFixedErnieScore & \RouterBiFixedErnieScore & \RouterBiFixedErnieCost & \RouterBiFixedErnieSaveErnie/\RouterBiFixedErnieSaveFlux\% \\
FLUX.2-dev & \RouterEnFixedFluxScore & \RouterZhFixedFluxScore & \RouterBiFixedFluxScore & \RouterBiFixedFluxCost & \RouterBiFixedFluxSaveErnie/\RouterBiFixedFluxSaveFlux\% \\
Uniform random & \RouterEnRandomScore & \RouterZhRandomScore & \RouterBiRandomScore & \RouterBiRandomCost & \RouterBiRandomSaveErnie/\RouterBiRandomSaveFlux\% \\
Q-Profile-Q & \textbf{\RouterEnProfileQScore} & \textbf{\RouterZhProfileQScore} & \textbf{\RouterBiProfileQScore} & \RouterBiProfileQCost & \RouterBiProfileQSaveErnie/\RouterBiProfileQSaveFlux\% \\
Q-Profile-C & \RouterEnProfileCScore & \RouterZhProfileCScore & \RouterBiProfileCScore & \textbf{\RouterBiProfileCCost} & \textbf{\RouterBiProfileCSaveErnie}/\textbf{\RouterBiProfileCSaveFlux}\% \\
\bottomrule
\end{tabular}
\caption{Out-of-fold routing under HCQ. Bi. is the fixed EN/ZH average; Save
E/F is relative to fixed ERNIE-Image/FLUX2-dev. Q and C denote quality-first
and cost-aware routing.}
\label{tab:routing_cost}
\end{table}

\paragraph{Quality-cost results.}
Q-Profile-Q gives the highest routed point estimate, but its paired gain over
ERNIE is marginal at the lower endpoint and comes at higher cost. Q-Profile-C
matches ERNIE's point estimate while reducing cost by 21.3\% relative to ERNIE
and 65.0\% relative to FLUX2. Its interval against ERNIE crosses the preregistered 0.20-point
non-inferiority margin, so the result supports a quality--cost tradeoff rather
than lossless routing. Uniform random performs worse, excluding arbitrary model
assignment as the explanation.

\section{Conclusion}
QC-T2I-Bench turns question decomposition from an intermediate step toward a
prompt score into reusable evidence. Its attributed records combine five
capability groups, dependency validity, and DSG context: HCQ yields
complexity-aware rankings, while graph controls separate root realization from
attached requirements. Across 13 models and two languages, exact completion
falls from 80.7\% for two-tag components to 37.2\% for components with seven or
more tags; after marginal control, excess coupling is resolved on direct DSG
edges but not matched non-ancestral pairs, localizing rather than causally
identifying composition-associated dependence. The same records support a
training-free router that matches ERNIE's point estimate with 21.3\% lower
GPU-s/MP, although its interval precludes a lossless-routing claim. Thus one
auditable evidence base supports ranking, diagnosis, and cost-aware selection
without treating any scalar score as universal. Future work should calibrate
question construction, DSG parsing, and adjudication across domains.

\clearpage{}
\bibliography{references}
\end{document}